\documentclass[10pt,twocolumn,letterpaper]{article}

\usepackage{wacv}
\usepackage{times}
\usepackage{epsfig}
\usepackage{graphicx}
\usepackage{amsmath}
\usepackage{amssymb}
\usepackage{pifont}
\usepackage{booktabs}
\usepackage{multirow}
\usepackage{color}
\def\wacvPaperID{399} % Enter the WACV Paper ID here

\wacvapplicationstrack % Uncomment this line if you are suboldsymbolitting to the Applications Track.

\wacvfinalcopy % *** Uncomment this line for the final suboldsymbolission

\ifwacvfinal
\usepackage[breaklinks=true,bookmarks=false]{hyperref}
\else
\usepackage[pagebackref=true,breaklinks=true,colorlinks,bookmarks=false]{hyperref}
\fi

\begin{document}

%%%%%%%%% TITLE
%\title{RoSTFine: Role-Separated Transformer \\ for Fine-grained and Diverse Sperm Feature Extraction}
%\title{A Framework for Automated Sperm Assessment and\\ A Neural Network Specialized for Sperm Video Recognition}
\title{Automated Sperm Assessment Framework and\\Neural Network Specialized for Sperm Video Recognition}

\author{Takuro Fujii${}^{1}$\hspace{6mm}Hayato Nakagawa${}^{1}$\hspace{6mm}Teppei Takeshima${}^{2}$\hspace{6mm}Yasushi Yumura${}^{2}$\hspace{6mm}Tomoki Hamagami${}^{1}$
\vspace{2mm}\\
${}^{1}$Yokohama National University\hspace{6mm}${}^{2}$Yokohama City University Medical Center\\
{\tt \small\{tkr.fujii.ynu, haya.nakagawa.ynu, hamagami.ynu\}@gmail.com}
\hspace{3mm}
{\tt \small\{yumura, teppei\_t\}@yokohama-cu.ac.jp}
}

\maketitle
\thispagestyle{empty}

%%%%%%%%% ABSTRACT
\begin{abstract}

%Infertility is a global health problem and more and more couples are seeking medical assistance to achieve reproduction, at least half of which are caused by men.
%In infertility treatment, sperm assessments and selection are performed manually which is dependent on the skills of the experts and as such lead to human errors and inconsistent assessments results among individuals and clinics. 
%Machine learning is expected to enable consistent, automated, and expedited sperm assessments.
%In this work, we construct the sperm video dataset and propose the neural network, RoSTFine, to assess sperms. RoSTFine captures fine-grained and diverse sperm morphology and motility from videos by selecting important parts and modeling shape and motion capturing process separately.
%Experimental results show that our proposed network improves the performance compared to many existing models. More analysis show that RoSTFine can attend strongly to sperm head and neck which are important parts for assessments.

Infertility is a global health problem, and an increasing number of couples are seeking medical assistance to achieve reproduction, at least half of which are caused by men.
The success rate of assisted reproductive technologies depends on sperm assessment, in which experts determine whether sperm can be used for reproduction based on morphology and motility of sperm.
Previous sperm assessment studies with deep learning have used datasets comprising images that include only sperm heads, which cannot consider motility and other morphologies of sperm.
Furthermore, the labels of the dataset are one-hot, which provides insufficient support for experts, because assessment results are inconsistent between experts, and they have no absolute answer.
Therefore, we constructed the video dataset for sperm assessment whose videos include sperm head as well as neck and tail, and its labels were annotated with soft-label.
Furthermore, we proposed the sperm assessment framework and the neural network, RoSTFine, for sperm video recognition.
Experimental results showed that RoSTFine could improve the sperm assessment performances compared to existing video recognition models and focus strongly on important sperm parts ({\it i.e.}, head and neck).
% 研究の意義
%We addressed reproduction, an important medical issue in human life but little-studied in computer vision fields, and our study has the potential to make a contribution to human well-being.
Our code is publickly available at \url{https://github.com/FTKR12/RoSTFine} .

\end{abstract}

\vspace{-4mm}
\section{Introduction}

% 一般大衆向け背景
Infertility is a critical problem around the world. This afflicts one in six couples, at least half of whom are casued by men~\cite{ISIDORI2005314,10.1093/humrep/dei307}.
Assisted reproductive technologies (ARTs), such as in-vitro-fertilization (IVF) and intracytoplasmic sperm injection (ICSI), are used depending on the cause and severity of infertility.
However, ARTs are currently successful in only approximately 33\% of cases, and this main reason is suboptimal sperm selection~\cite{PMID:29089604}.
In the sperm selection process, at least three fertility factors are typically examined; sperm concentration, motility and morphology~\cite{Kumar2015ProvidingHS}.
In sperm selection, motility and sperm concentration are assessed using computer-aided semen analysis (CASA) systems, which are sensitive to sample preparation and equipment setup~\cite{PMID:31406675,AMANN20145}.
Morphology is assessed manually by experts, which are inconsistent among individuals and clinics owing to subjective criteria, in addition to being time-consuming and labor-intensive~\cite{PMID:28692759,DAVIS1992763,PMID:2254406,PMID:2394790}.
%In sperm selection, motility and sperm concentration are assessed using computer-aided semen analysis (CASA) systems, whereas morphology is assessed manually by experts.
%CASA systems are sensitive to sample preparation and equipment setup~\cite{PMID:31406675,AMANN20145}, and manual assessments are inconsistent among individuals and clinics owing to subjective criteria, in addition to being time-consuming and labor-intensive~\cite{PMID:28692759,DAVIS1992763,PMID:2254406,PMID:2394790}.
%Therefore, an End2End system from sperm evaluations to sperm selections is in high demand and very promising for improving reproductive success.
Therefore, an End2End sperm assessment framework, considering all three factors, is in high demand and promising for improving reproductive success.

\begin{figure}
\includegraphics[width=8.3cm]{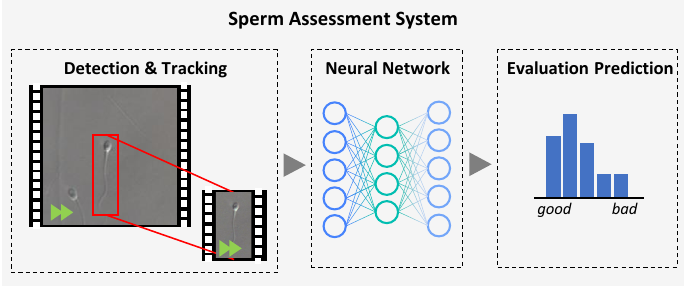}
   \caption{Proposed sperm assessment framework whose inputs are sperm videos. It detects and tracks a target sperm in a video taken from a microscope to make an input of a neural network, and then the neural network predicts the \textit{grade distribution} of the sperm.}
\label{fig:overview}
\vspace{-3mm}
\end{figure}

% 機械学習用いるためのモチベと既存研究の課題
Deep learning has shown promise for standardizing and automating sperm assessments~\cite{PMID:34002070,PMID:33581832,PMID:28834185}.
Several studies have addressed sperm assessment using deep learning~\cite{RIORDON2019103342,8602272,Spencer2022EnsembledDL,ILHAN2020103845,Yüzkat}. However, there are two problems. First, at the viewpoint of an End2End sperm assessment framework, despite the fact that the head morphology as well as other morphologies and motility are important factors in sperm selection~\cite{10665-343208}, these studies assessed sperm by classifying only head morphology, ignoring neck and tail morphology and sperm motility.
Second, at the viewpoint of support for experts, the annotation of the data used in these studies was a one-hot label for classification tasks.
As the labels of a dataset for sperm assessment, soft-labels are better because there is no absolute answer and experts have inconsistent assessments. Soft-labels enable to aid experts in flexible decision-making because soft-labels are informative.
%Riordon \etal \cite{RIORDON2019103342} assesses sperm by classifying sperm head morphology using the VGG16 based model.
%Prabaharan \etal \cite{8602272} proposes the original CNN model for anomaly detection of sperm head morphology. Others and details are given in \S\ref{sec:related work}.
%Despite the fact that not only the head morphology but also other morphology and motility are important factors in sperm selection~\cite{10665-343208}, most of the previous researches are about sperm assessments by classifying only head morphology, ignoring neck and tail morphology and sperm motility.
%In addition, the labels for the data used in these works are hard labels, which are ethically problematic at the viewpoint of practical applications involving human life.

% 精子認識上の課題
For video application tasks, it is straightforward approach to apply existing video recognition models directly.
Their models, however, are designed for general videos whose domain is different from sperm videos.
Therefore, we should design a model specialized for sperm video recognition, but there are two challenges. First, it is difficult for a model to capture only a target sperm because videos are background dominant, and dust, air bubbles and non-target sperm can interfere with sperm recognition.
Second, the model must capture the diverse characteristics of sperm, such as motility, morphologies and dependencies of the head, neck and tail.
It is particularly difficult to capture the sperm tail, which is often assimilated into the background.

% 取り組む内容
To solve these problems, in this study, we constructed a video dataset annotated with soft-labels, and proposed an End2End framework for sperm assessment and a neural network for sperm video recognition.
%In this paper, to address these problems, we constructed the video dataset for sperm assessments, and proposed the novel sperm assessment system and neural network.
When constructing the dataset, each of the 40 experts annotated one of the five grades for each sample, thus, the labels are soft-labels, 5-grade histograms, which we refer to as \textit{grade distribution}.
%The dataset are annotated in five grades by about 40 experts, whose labels are a distribution of 5 grade classes.
%We refer to this soft-label as \textit{grade distribution}.
The details of the dataset are presented in \S\ref{sec:dataset}.
The proposed framework, illustrated in Figure \ref{fig:overview}, detects and tracks a target sperm in a video captured from a microscope to provide an input to a neural network, and then predicts the \textit{grade distribution} of the sperm.
%The proposed sperm assessment system detects and tracks a target sperm in a video taken from a microscope to make an input of a neural network, and then it predicts the sperm evaluation.
The proposed neural network, \textit{Role-Separated Transformer  for Fine-Grained and Diverse Sperm Feature Extraction} (RoSTFine), can focus only on a target sperm and extract fine-grained and diverse sperm features.
The details of RoSTFine are presented in \S\ref{sec:method}.
The experimental results (\S\ref{sec:experiments}) show that RoSTFine achieves a higher performance than existing video recognition models, such as TimeSformer and SlowFast~\cite{Feichtenhofer_2019_ICCV}. 
Further Analysis showed that RoSTFine can attend strongly to the sperm head and neck which are important for sperm assessment, and can generate effective features. 

% 貢献
Our main contributions are summarized as follows: 
(1) To address reproduction that is an important issue but little-studied in computer vision fields, we constructed a video dataset annotated with soft-labels for sperm assessment;
(2) We developed an automated framework for sperm assessment;
(3) We developed a sperm-specific model, RoSTFine, to capture important sperm characteristics;
(4) Experimental results showed that RoSTFine improved assessment performances on three evaluation metrics;
(5) RoSTFine can focus on important sperm parts, such as the head and neck.
% \begin{itemize}
%  \item To address reproduction that is an important issue but little-studied in computer vision fields, we constructed a video dataset for sperm assessment.
%  \item We developed an automated framework for sperm assessment using videos as inputs.
%  \item We developed a sperm-specific transformer model, RoSTFine, to capture important sperm characteristics and improve sperm assessment performance compared to existing video recognition models.
% \end{itemize}

%------------------------------------------------------------------------
\section{Related Work} \label{sec:related work}

\subsection{Datasets and Methods for Sperm Assessment}
There are three publicly available datasets for sperm assessment.
The Sperm Morphology Image Dataset (SMIDS)~\cite{article} comprises 3000 images of sperm head, and is annotated in three classes of normal, abnormal, and non-sperm.
The human sperm head morphology (HuSHeM) dataset~\cite{SHAKER2017181} comprises 216 images of stained sperm head, and is annotated in four classes of normal, tapered, pyriform, and amorphous. 
The Laboratory for Scientific Image Analysis Gold-standard for Morphological Sperm Analysis (SCIAN) dataset~\cite{CHANG2017143} comprises 1132 images of sperm head, and is annotated in five classes of small and the same 4 classes as the HuSHeM dataset.
The classes of the HuSHeM and SCIAN dataset are subsets of the categories of sperm head morphology which World Health Organization (WHO) provided in 
the semen analysis manual~\cite{10665-343208}.

%Some studies have used deep learning methods for these datasets.
In some studies, these datasets have been used to train deep learning models.
Riordon \etal~\cite{RIORDON2019103342} fine-tuned VGG16 pretrained on ImageNet to classify sperm 
head morphology.
Spencer \etal~\cite{Spencer2022EnsembledDL} used a stacked ensemble comprising VGG16, VGG19, ResNet-34, and DenseNet-161.
Yüzkat \etal~\cite{Yüzkat} designed and fusioned six CNN-based models.
Ilhan \etal~\cite{ILHAN2020103845} proposed a computational framework that includes multistage cascade-connected preprocessing techniques, region-based descriptor features, and nonlinear kernel SVM-based learning.

However, these datasets and studies are incomplete for sperm assessment because they focus only on sperm head morphology, although other morphologies and motility are also important.
Additionally, although there are no absolute answer and assessment results are sometimes inconsistent, the labels of the datasets are one-hot labels.
In this study, we constructed a sperm video dataset annotated with soft-labels, and developed a sperm recognition model to capture sperm morphology and motility.

\subsection{Video Recognition Models}
Video recognition is one of the most popular computer vision fields, and many neural networks have been developed. Video recognition models are classified into two categories: CNN- and Transformer-based models.
In CNN-based models, Two-Stream I3D~\cite{Carreira_2017_CVPR} has an RGB stream and an optical flow stream, and its backbone model is a 3D convolutional network.
SlowFast~\cite{Feichtenhofer_2019_ICCV} involves a slow pathway to capture shapes and slow motion and a fast pathway to capture fast motion and movement. 
In Transformer-based models, ViViT~\cite{Arnab_2021_ICCV} combines spatial and temporal attention in various ways. TimeSformer~\cite{pmlr-v139-bertasius21a} proposes various methods for calculating temporal attention. Althogh CNN-based models can achieve high performance even with small datasets, owing to their strong inductive bias, they have a narrow receptive field and are poor at capturing long dependencies~\cite{Yamins2014PerformanceoptimizedHM}.
Transformer-based models require large dataset for high performance owing to their weak inductive bias, however, they have a wide receptive field and can capture long dependencies~\cite{Tuli2021AreCN}.
Furthermore, the model structures are flexible and easy to operate, making them useful not only in computer vision but also in various fields, such as natural language processing~\cite{devlin-etal-2019-bert}, speech recognition~\cite{Chang2021}, and multi-modal processing~\cite{pmlr-v139-radford21a}. 

These models achieve high performances in some benchmark datasets, such as Kinetics~\cite{Carreira_2017_CVPR}, Diving-48~\cite{Li_2018_ECCV}, and Something-Something-V2~\cite{goyal2017something}, but these models and datasets have been developed for general video recognition.
However, sperm videos are different from general videos because they are captured using a microscope.
Therefore, sperm-specific models must be developed.
In~\cite{RIORDON2019103342, Spencer2022EnsembledDL}, the effectiveness of fine-tuning models pretrained on general images in sperm image recognition was shown. Inspired by this and the high operability of local features, we developed Transformer-based model to utilize a pretrained model.

%------------------------------------------------------------------------
\section{Dataset and Task Definition} \label{sec:dataset}

We constructed a sperm video dataset for sperm assessment.
When constructing the dataset, each of the 40 experts annotated one of the five grades for each sample.
The grades are as follows: A (best); B (good); C (neither); D (bad); and E (worst).
To replicate the actual variability, the experts graded them based on their knowledge and senses.
Therefore, the label of the dataset is a soft-label, which is a 5-grade histogram. We refer to this soft-label as the \textit{grade distribution}.
The dataset includes 615 videos captured using a microscope. 
The videos are 175-frame clips at 15 frames per second with 1392 × 976-pixel crops.
%The grade distribution is normalized by dividing it by the number of experts, such that the sum is 1.

\begin{figure}[t]
\centering
\includegraphics[width=8.3cm]{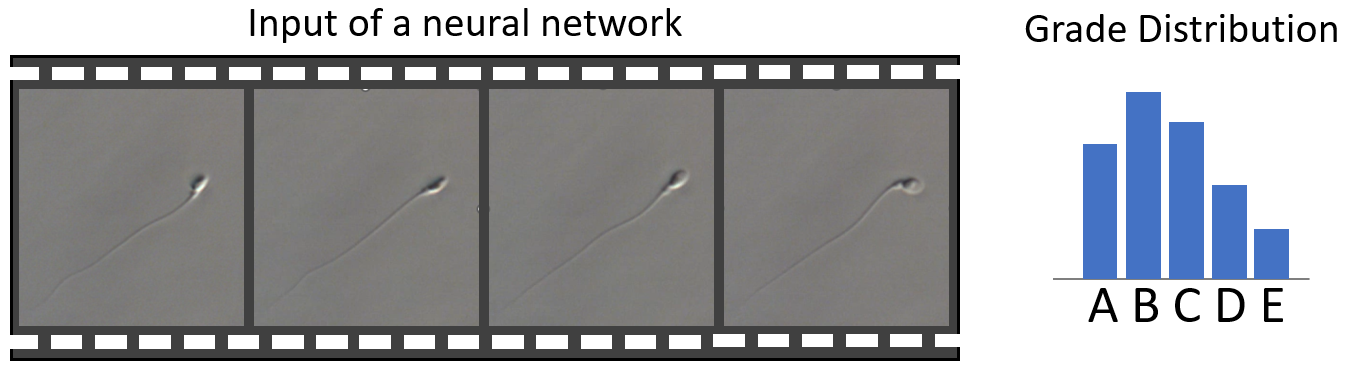}
   \caption{Sample tracking video and label of the dataset.}
\label{fig:video_label}
\end{figure}

We applied sperm detection and tracking to all videos to create inputs for the neural network because they were taken from a microscope, 
the target sperm was considerably small, and debris, such as air bubbles and other sperm, were reflected. This preprocessing is shown on the left side of Figure \ref{fig:overview}.
We tagged the target sperm in the first frame of all videos, and tracked it by template matching to detect and track sperm.
The tracked videos are 16-frame clips with 150 × 150-pixel crops.
The dataset statistics is shown in Table \ref{tab:data}, and a sample of the tracking videos and labels are shown in Figure \ref{fig:video_label}. More samples, including original videos, tracking videos, and labels, are shown in Figure A.2.

We define the sperm assessment task in our dataset as the 5-point regression task, because information of the most selected grade as well as that of the other grades are important for decision support.
Given an input video $\mathcal{V}\in\mathbb{R}^{H\times W\times 3\times T}$, a neural network predicts the \textit{grade distribution} $\hat{Y}\in\mathbb{R}^{5}$.

\section{Method} \label{sec:method}
\subsection{Method Overview}
% 既存手法の概要と課題（グローバル特徴のみでタスク解くという課題・精子での課題
Transformer models capture global dependencies through self-attention, and have achieved high performances in various vision tasks.
However, self-attention treats each local patch uniformly to calculate the attention score, and then computes a weighted sum of all local patches.
% 図2のサンプルもうちょい寄せで
\begin{table}[t]
  \begin{center}
\begin{tabular}{c|ccccc|c}
\toprule
\multirow{2}{*}{\textbf{train / test}} & \multicolumn{5}{|c|}{\textbf{Grade}} & \multirow{2}{*}{\textbf{Total}}\\
 & \textbf{A} & \textbf{B} & \textbf{C} & \textbf{D} & \textbf{E} & \\
\midrule
492 / 123 & 45 & 194 & 356 & 9 & 11 & 615 \\
\bottomrule
\end{tabular}
\end{center}
\caption{Statistics of the dataset. Each value of Grade is the number of the samples for which the most experts selected the grade.}
\label{tab:data}
\end{table}
A global feature is dominated by all local patches, thus, simultaneously considering all local patches may reduce the influence of some important local patches.
Particularly, in a sperm recognition task that require the capture of fine-grained shapes and motions, this method may cause serious  discriminative deficiencies.
Therefore, we propose \textit{Patch Selection Module} (PSM) to select only important and informative patches. 

% 精子認識タスク依存の重要な課題（形状、動きなど様々な特徴が必要）
% 精子認識タスクにおいてもうひとつの解決すべき課題として、多様な特徴抽出が必要である点がある。
Another challenge in sperm recognition is the extraction of diverse sperm features, such as the morphologies, motions and dependencies of various sperm parts.
We expect to extract diverse features using global and local features effectively.
We propose \textit{Role-Separated Branch} (RSB) to effectively use local patches obtained by PSM and extract the spatial and temporal features separately.

\begin{figure*}
\begin{center}
\includegraphics[width=17.4cm]{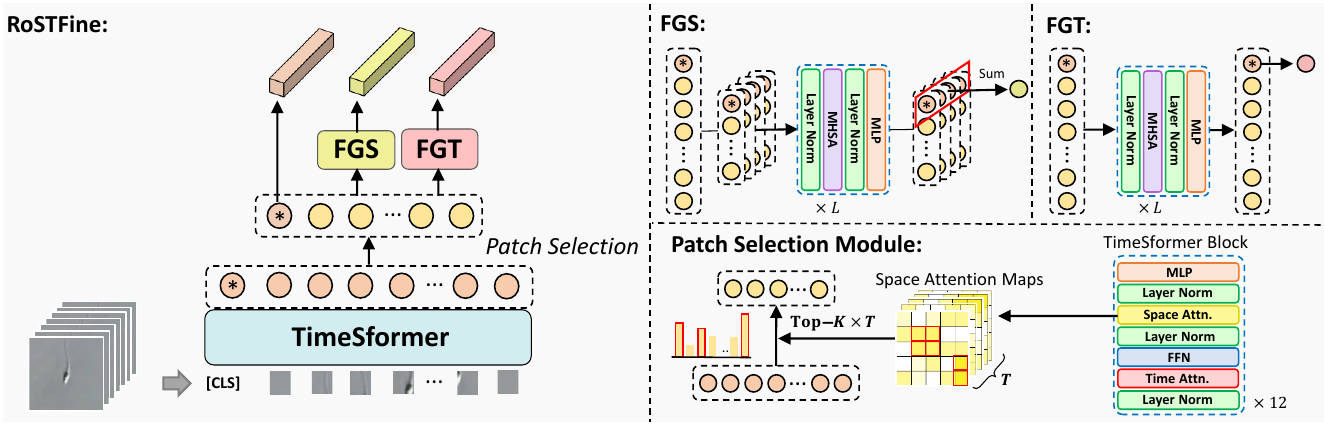}
\end{center}
   \caption{Architecture of RoSTFine. The video encoder of RoSTFine is TimeSformer. \textit{Patch Selection Module} (PSM) selects important and informative local patches based on attention maps obtained from TimeSformer. \textit{Fine-Grained Spatial Feature Extraction Branch} (FGS) applies multi-head self attention to each frame. \textit{Fine-Grained Temporal Feature Extraction Branch} (FGT) applies multi-head self attention across all frames.}
\label{fig:model}
\end{figure*}

We propose \textit{Role-Separated Transformer for Fine-Grained and Diverse Sperm Feature Extraction} (RoSTFine), putting PSM and RSB on the head of TimeSformer. The details of RoSTFine are described below and its architecture is illustrated in Figure \ref{fig:model}.

%We propose RoSTFine consisting of TimeSformer, PSM and RSB.

% motivation and overview
\subsection{Video Encoder}
TimeSformer~\cite{pmlr-v139-bertasius21a} is used as the sperm video encoder.
An input video $\boldsymbol{X} \in \mathbb{R}^{H\times W\times 3\times T}$ comprising $T$ RGB frames of size $H\times W$ is decomposed into $N$ non-overlapping patches for each frame, each of size $P\times P$. Each patch $\boldsymbol{x}_{(t,p)}\in\mathbb{R}^{3P^2}$ is linearly mapped into an embedding $\boldsymbol{z}_{(t,p)}\in\mathbb{R}^{d}$ using a learnable matrix $\boldsymbol{E}\in\mathbb{R}^{d\times3P^2}$.
\begin{equation}
    \small
    \boldsymbol{z}_{(t,p)} = \boldsymbol{E}\boldsymbol{x}_{(t,p)} + \boldsymbol{e}_{(t,p)}
\end{equation}
where $\boldsymbol{e}_{(t,p)}$ denotes the learnable positional embeddings added to encode the spatiotemporal position of each patch.
The input of TimeSformer is represented by Eq. \ref{input}. 
\begin{equation}
    \small
    \boldsymbol{Z}=\{\boldsymbol{z}_{cls},\boldsymbol{z}_{(1,1)},...,\boldsymbol{z}_{(T,N)}\}\in\mathbb{R}^{(1+NT)\times d}
    \label{input}
\end{equation}
where $\boldsymbol{z}_{cls}$ is a learnable \texttt{[CLS]} embedding.
The input $\boldsymbol{Z}$ is fed into TimeSformer blocks, which comprises a spatial attention, a temporal attention, and a multi-layer-perceptron.
The output of TimeSformer is represented by Eq. \ref{output}.
\begin{equation}
    \small
    \boldsymbol{V}=\{\boldsymbol{v}_{cls},\boldsymbol{v}_{(1,1)},...,\boldsymbol{v}_{(T,N)}\} \in \mathbb{R}^{(1+NT)\times d}
    \label{output}
\end{equation}

\subsection{Patch Selection Module}
% 一般的な説明→課題→そのためtoken selection moduleでは、→理論→式
Transformer models solve various tasks using \texttt{[CLS]} embedding aggregated by all local patches. Using  only the \texttt{[CLS]} embedding may cause serious discriminative deficiencies, especially in a sperm recognition task, whose data are background dominant and which requires some fine-grained parts. Certain critical local patches have the potential to capture fine-grained sperm features.

We propose \textit{Patch Selection Module} (PSM) to select important local patches based on attention scores. Specifically, the spatial attention of each TimeSformer block generates an attention map $\boldsymbol{A}^l \in \mathbb{R}^{T\times(1+N)\times(1+N)}$, which represents the correlation with all patches. Here, $l$ is the number of layers and ranges from $1$ to $L$. To select patches, PSM uses the attention scores $\boldsymbol{A}^*$ calculated as the sum of the attention scores of \texttt{[CLS]} of the last two layers.
\begin{equation}
    \small
    \boldsymbol{A}^{*} = \boldsymbol{A}^{L-1}[:,0,1:] + \boldsymbol{A}^{L}[:,0,1:] = \{\boldsymbol{A}_1^*,..., \boldsymbol{A}_T^*\}
    \in\mathbb{R}^{T\times N\times N}
\end{equation}
The top $K$ patches are selected in $N$ patches in each frame corresponding to the top $K$ highest scores in $\boldsymbol{A}_i^*$. 
The selected patches are represented by Eq. \ref{select}.
\begin{equation}
    \small
    \boldsymbol{F}=\{\boldsymbol{f}_{(1,1)},\boldsymbol{f}_{(1,2)},...,\boldsymbol{f}_{(T,K)}\} \in \mathbb{R}^{TK\times d}
    \label{select}
\end{equation}
where $\boldsymbol{f}_{(t,k)}$ denotes the embeding of the $k$-patch in $t$-frame.
The process of PSM is shown in the bottom right of Figure \ref{fig:model}.

\subsection{Role-Separated Branch}
% 課題・概要・役割→簡単な説明
In a sperm recognition task, it is important to capture diverse sperm features, such as shapes, motions and their dependencies. However, \texttt{[CLS]} embedding may miss these features, because it is calculated using all local patches in the same manner and includes a large background. Therefore, we propose \textit{Role-Separated Branch} (RSB) to effectively use the local patches and extract fine-grained spatial and temporal features separately and explicitly, using the local patches selected in PSM.
RSB comprises \textit{Fine-Grained Spatial Feature Extraction Branch} (FGS) and \textit{Fine-Grained Temporal Feature Extraction Branch} (FGT). 

%Many video recognition models are composed of multiple streams such as the spatial, temporal and optical flow stream. 
%Inspired by these models, we propose Role-Separated Branch which extracts the fine-grained spatial and temporal features separately from the local tokens selected in Token Selection Module.
%The structure of explicit role separation will facilitate learning.
% 一般的な動画像認識モデル（引用5つくらい）では、オプティカルフローや空間特徴および時空間特徴などの複数のストリームをもつモデルが多い。
% これらからヒントを得て、token selection moduleで選択したトークンを用いて細粒度な形状特徴・動き特を徴ブランチによって別々に抽出するRole-Separated Branchを提案する。
% 具体的には、フレーム内でアテンションを取る細粒度空間特徴抽出ブランチと、フレーム横断的にアテンションを取る細粒度動き特徴抽出ブランチによって構成される
% ブランチによって役割を明示的に分けることで学習の促進が期待される。
% 各ブランチの詳細な説明は以下で述べる。

\textit{Fine-Grained Spatial Feature Extraction Branch} \label{fgs}
% 簡単な説明・理論・式
% FGSモジュールは、選択された重要なトークンのみを用いてフレーム内でアテンションをとることによって、細粒度な形状特徴を抽出します。
(FGS) obtains spatial sperm features, such as shape, texture, and dependencies within a frame.
The inputs $\boldsymbol{G}^{s,(0)}$ of FGS are obtained by dividing $\boldsymbol{F}$ into frame units, and then attaching \texttt{[CLS]} embedding to each unit. The $i$-frame unit is represented by Eq. \ref{eq:inputoffgs}. 
\begin{equation}
    \small
    \boldsymbol{G}^{s,(0)}_i = \{\boldsymbol{v}_{cls},\boldsymbol{g}_{(i,1)}^s,...,\boldsymbol{g}_{(i,K)}^s\}\in\mathbb{R}^{(1+K)\times d}
    \label{eq:inputoffgs}
\end{equation}
Each $\boldsymbol{G}^{s,(0)}_i$ is fed into $L$ Attention blocks, and 
then we obtain the output $\boldsymbol{G}^{s,(L)}_i$ (Eq. \ref{eq:fgs}).
\begin{equation}
    \small
    \boldsymbol{G}_i^{s,(l)}=\text{MLP}(\text{MHSA}(\text{LN}(\boldsymbol{G}_i^{s,(l-1)})))
    \label{eq:fgs}
\end{equation}
where LN, MHSA and MLP denote Layer Normalization, Multi-Head Self-Attention, and Multi-Layer-Perceptron, respectively, and $l$ represents the number of attention blocks.
Finally, the fine-grained spatial feature $\boldsymbol{v}^s$ is obtained from the mean of all \texttt{[CLS]} embeddings $\boldsymbol{G}_i^{s,(L)}[0]\hspace{1mm}(i=1,...,F)$ of the last layer (Eq. \ref{eq:fgs2}).
\begin{equation}
    \small
    \boldsymbol{v}^s=\frac{1}{KT}\sum_{i=1}^F \boldsymbol{G}_i^{s,(L)}[0]
    \label{eq:fgs2}
\end{equation}
The FGS process is shown in 
the top right of Figure \ref{fig:model}.

\textit{Fine-Grained Temporal Feature Extraction Branch}
(FGT) obtains temporal features ({\it e.g.}, motions, movements, and dependencies between frames).
The input $\boldsymbol{G}^{t,(0)}$ of FGT is obtained by attaching \texttt{[CLS]} embedding $\boldsymbol{v}_{cls}$ to $\boldsymbol{F}$.
\begin{equation}
    \small
    \boldsymbol{G}^{t,(0)}=\{\boldsymbol{v}_{cls}, \boldsymbol{f}_1, \boldsymbol{f}_2,...,\boldsymbol{f}_{K\times T}\}
    \in \mathbb{R}^{(1+KT)\times d}
\end{equation}
$\boldsymbol{G}^{t,(0)}$ is fed into $L$ Attention blocks, and then we obtain the output $\boldsymbol{G}^{t,(L)}$ (Eq. \ref{eq:fgt}).
Finally, we obtain the \texttt{[CLS]} embedding as the fine-grained temporal feature $\boldsymbol{v}^t$ (Eq. \ref{eq:fgt2}).
\begin{equation}
    \small
    \boldsymbol{G}^{t,(l)}=\text{MLP}(\text{LN}(\text{MHSA}(\text{LN}(\boldsymbol{G}^{t,(l-1)}))))
    \label{eq:fgt}
\end{equation}
\begin{equation}
    \small
    \boldsymbol{v}^t=\boldsymbol{G}^{t,(L)}[0]
    \label{eq:fgt2}
\end{equation}
where $l$ denotes the number of the attention blocks.
The FGT process is shown in the top right of Figure \ref{fig:model}.
%------------------------------------------------------------------------

\subsection{Training and Inference} \label{training and inference}
\noindent\textbf{Training.}\hspace{1mm}
We regard the estimation of the \textit{grade distribution} of the sperm as a 5-point regression task because the \textit{grade distribution} is a 5-point histogram.
%We regard the sperm assessment grade estimation task as a 5-point regression task.
We use Mean Squared Error (MSE) as the training objective, represented by Eq. \ref{eq:MSE}.  
\begin{equation}
    \small
    \text{MSE}(\hat{\boldsymbol{y}},\boldsymbol{y}) = \frac{1}{n} \sum_{i=1}^n (\hat{\boldsymbol{y}}_i-\boldsymbol{y}_i)^2
    \label{eq:MSE}
\end{equation}
where $n$ denotes the number of data points, and $n=5$ in this case.
The training objective $\mathcal{L}_{mse}$ is calculated as the mean of all MSE of $\hat{\boldsymbol{y}}^g$, $\hat{\boldsymbol{y}}^s$, and $\hat{\boldsymbol{y}}^t$ (Eq. \ref{eq:MSE_}).
\begin{equation}
    \small
    \mathcal{L}_{mse} = \left(\text{MSE}(\hat{\boldsymbol{y}^g},\boldsymbol{y}) + \text{MSE}(\hat{\boldsymbol{y}^s},\boldsymbol{y}) + \text{MSE}(\hat{\boldsymbol{y}^t},\boldsymbol{y})\right)/3
    \label{eq:MSE_}
\end{equation}
where $\hat{\boldsymbol{y}}^g$, $\hat{\boldsymbol{y}}^s$, and $\hat{\boldsymbol{y}}^t$ are the predicted \textit{grade distributions} obtained by linear projection of $\boldsymbol{v}^g$, $\boldsymbol{v}^s$, and $\boldsymbol{v}^t$, respectively.

However, the \textit{grade distribution} estimation task can be regarded as a distribution distance minimizing problem.
Therefore, we also use JS-divergence as the training objective. JS-divergence is represented by Eq. \ref{eq:js}, which is a variant of KL-divergence, represented by Eq. \ref{eq:kl}, to smooth out the divergence and maintain symmetry.
\begin{equation}
    \small
    \text{KL}(P,Q)=\sum_i p_i\log\left(\frac{p_i}{q_i}\right)
    \label{eq:kl}
\end{equation}
\begin{equation}
    \small
    \text{JS}(P,Q)=\frac{1}{2}\text{KL}\left(P,\frac{P+Q}{2}\right)+\frac{1}{2}\text{KL}\left(Q,\frac{P+Q}{2}\right)
    \label{eq:js}
\end{equation}
The training objective $\mathcal{L}_{js}$ is calculated by the mean of all JS-divergence of $\hat{\boldsymbol{y}}^g$, $\hat{\boldsymbol{y}}^s$, and $\hat{\boldsymbol{y}}^t$ (Eq. \ref{eq:JS_}).
\begin{equation}
    \small
    \mathcal{L}_{js} = \left(\text{JS}(\hat{\boldsymbol{y}^g},\boldsymbol{y}) + \text{JS}(\hat{\boldsymbol{y}^s},\boldsymbol{y}) + \text{JS}(\hat{\boldsymbol{y}^t},\boldsymbol{y})\right)/3
    \label{eq:JS_}
\end{equation}

\noindent\textbf{Optional.}\hspace{1mm}
In addition, we use the diversity loss as an optional training objective. 
$\boldsymbol{v}^g$, $\boldsymbol{v}^s$, and $\boldsymbol{v}^t$ may be similar vectors because their vectors are generated from the same \texttt{[CLS]} embedding. To diversify them, we apply the orthogonal constraint (Eq. \ref{eq:divloss}) to any pairs of $\boldsymbol{v}^g$, $\boldsymbol{v}^s$ and $\boldsymbol{v}^t$, represented by Eq. \ref{eq:divloss_}. This constraint enables the cosine similarity between two vectors to be brought close to zero, and effectively facilitates them becoming different vectors.
\begin{equation}
    \small
    \text{div}(\boldsymbol{v}^i,\boldsymbol{v}^j) = \left| \frac{\boldsymbol{v}^i\cdot \boldsymbol{v}^j}{||\boldsymbol{v}^i||_2||\boldsymbol{v}^j||_2}\right|
    \label{eq:divloss}
\end{equation}
\begin{equation}
    \small
    \mathcal{L}_{div} = \left(\text{div}(\boldsymbol{v}^g,\boldsymbol{v}^s) + \text{div}(\boldsymbol{v}^g,\boldsymbol{v}^t) + \text{div}(\boldsymbol{v}^s,\boldsymbol{v}^t)\right)/3
    \label{eq:divloss_}
\end{equation}
We expect each vector to reclaim diverse features of sperm by the diversity loss $\mathcal{L}_{div}$. When applying the diversity loss, the training objective is represented as Eq. \ref{eq:train obj}
\begin{equation}
    \small
    \mathcal{L} = \mathcal{L}_{mse/js} + \alpha\mathcal{L}_{div}
    \label{eq:train obj}
\end{equation}
where $\alpha$ is the weight of the diversity loss.

\noindent\textbf{Inference.}\hspace{1mm}
The final predicted \textit{grade distribution} $\hat{y}$ is obtained from the mean of all predicted \textit{grade distributions} $\hat{\boldsymbol{y}}^g$, $\hat{\boldsymbol{y}}^s$, and $\hat{\boldsymbol{y}}^t$ (Eq. \ref{pred}).
\begin{equation}
    \small
    \hat{\boldsymbol{y}}= (\hat{\boldsymbol{y}}^g+\hat{\boldsymbol{y}}^s+\hat{\boldsymbol{y}}^t)/3
    \label{pred}
\end{equation}

The architecture of RoSTFine is expected to have an effect similar to that of ensemble learning because $\hat{\boldsymbol{y}}^g$, $\hat{\boldsymbol{y}}^s$, and $\hat{\boldsymbol{y}}^t$ are optimized, respectively. $\hat{\boldsymbol{y}}^g$ is optimized based on the entire sperm, $\hat{\boldsymbol{y}}^s$ is optimized based on the sperm shape, and $\hat{\boldsymbol{y}}^t$ is optimized based on the sperm motion and dependencies across frames.
We will show that this training and inference strategy is the best in \S\ref{ablation}.
%The RoSTFine architecture is expected to get the similar effect as an ensemble learning in addition to diverse feature extraction.

%------------------------------------------------------------------------
\section{Experiment} \label{sec:experiments}

\begin{table*}[h]
  \begin{center}
    {\small{
\begin{tabular}{lcc|cccccc}
\toprule
\multirow{2}{*}{\textbf{Method}} & \multirow{2}{*}{\textbf{MSE}$^{(10^{-2})}$} & \multirow{2}{*}{\textbf{JS div.}$^{(10^{-2})}$} & \multicolumn{6}{c}{\textbf{Balanced Accuracy (\%)}}\\
&&& \textbf{1st} & \textbf{2nd} & \textbf{3rd} & \textbf{4th} & \textbf{5th} & \textbf{Avg.}\\
\midrule
VGG16$^{\dag}$ & 1.517 \scriptsize{$\pm$ 0.09} & 5.628 \scriptsize{$\pm$ 0.27} & 26.49 \scriptsize{$\pm$ 3.40} & 22.72 \scriptsize{$\pm$ 4.44} & 22.31 \scriptsize{$\pm$ 5.11} & 22.25 \scriptsize{$\pm$ 4.12} & 33.44 \scriptsize{$\pm$ 6.59} & 25.44\\
\midrule
R3D & 1.365 \scriptsize{$\pm$ 0.10}& 4.978 \scriptsize{$\pm$ 0.38} & 28.72 \scriptsize{$\pm$ 4.16} & 31.03 \scriptsize{$\pm$ 2.50} & 29.70 \scriptsize{$\pm$ 4.30} & 28.46 \scriptsize{$\pm$ 6.35} & 34.68 \scriptsize{$\pm$ 3.29} & 30.52\\
R(2+1)D & 1.702 \scriptsize{$\pm$ 0.09} & 7.043 \scriptsize{$\pm$ 0.28} & 29.83 \scriptsize{$\pm$ 3.42} & 21.83 \scriptsize{$\pm$ 1.49} & 20.26 \scriptsize{$\pm$ 0.51} & 22.64 \scriptsize{$\pm$ 2.78} & 26.96 \scriptsize{$\pm$ 5.50} & 24.30\\
X3D & 1.808 \scriptsize{$\pm$ 0.07} & 7.186 \scriptsize{$\pm$ 0.23} & 27.65 \scriptsize{$\pm$ 4.33} & 20.46 \scriptsize{$\pm$ 0.99} & 22.33 \scriptsize{$\pm$ 1.67} & 22.74 \scriptsize{$\pm$ 2.45} & 32.12 \scriptsize{$\pm$ 7.03} & 25.06\\
I3D & 1.376 \scriptsize{$\pm$ 0.13} & 5.206 \scriptsize{$\pm$ 0.40} & 30.28 \scriptsize{$\pm$ 4.77} & 31.39 \scriptsize{$\pm$ 1.74} & 29.16 \scriptsize{$\pm$ 1.67} & 27.44 \scriptsize{$\pm$ 4.07} & 36.65 \scriptsize{$\pm$ 10.20} & 30.98\\
SlowFast & 1.346 \scriptsize{$\pm$ 0.13} & 5.059 \scriptsize{$\pm$ 0.54} & 31.58 \scriptsize{$\pm$ 5.36} & 30.00 \scriptsize{$\pm$ 2.41} & 29.15 \scriptsize{$\pm$ 4.46} & 24.85 \scriptsize{$\pm$ 5.06} & 34.65 \scriptsize{$\pm$ 8.03} & 30.05\\
ViViT & 1.406 \scriptsize{$\pm$ 0.09} & 4.987 \scriptsize{$\pm$ 0.39} & 27.37 \scriptsize{$\pm$ 0.73} & 28.34 \scriptsize{$\pm$ 3.28} & 28.26 \scriptsize{$\pm$ 4.65} & \textbf{31.46} \scriptsize{$\pm$ 4.55} & 39.50 \scriptsize{$\pm$ 5.70} & 30.99\\
TimeSformer & 1.186 \scriptsize{$\pm$ 0.10} & 4.283 \scriptsize{$\pm$ 0.17} & 31.43 \scriptsize{$\pm$ 1.32} & 35.62 \scriptsize{$\pm$ 3.72} & 32.49 \scriptsize{$\pm$ 4.05} & 28.47 \scriptsize{$\pm$ 5.19} & 41.68 \scriptsize{$\pm$ 5.46} & 33.94\\
\midrule
$\text{RoSTFine}_{\alpha=0}$ & \textbf{1.121} \scriptsize{$\pm$ 0.11} & \textbf{4.145} \scriptsize{$\pm$ 0.26} & \textbf{33.48} \scriptsize{$\pm$ 6.97} & 35.56 \scriptsize{$\pm$ 2.00} & \textbf{33.40} \scriptsize{$\pm$ 3.15} & 30.17 \scriptsize{$\pm$ 5.59} & \textbf{43.59} \scriptsize{$\pm$ 6.39} & \textbf{35.24}\\
$\text{RoSTFine}_{\alpha=1}$ & \textbf{1.104} \scriptsize{$\pm$ 0.12} & \textbf{4.109} \scriptsize{$\pm$ 0.26} & \textbf{35.61} \scriptsize{$\pm$ 4.19} & \textbf{36.71} \scriptsize{$\pm$ 3.55} & \textbf{35.97} \scriptsize{$\pm$ 5.51} & 30.29 \scriptsize{$\pm$ 6.10} & \textbf{43.29} \scriptsize{$\pm$ 8.58} & \textbf{36.37}\\
\bottomrule
\end{tabular}
}}
\end{center}
\caption{Comparing RoSTFine to the baselines. The value is average $\pm$ standard deviation of 5 folds. $\text{RoSTFine}_{\alpha=0,1}$ achieves the best performance in MSE, JS-divergence (JS div.) and most Balanced Accuracies (BA). $\dag$ denotes the lower bound, which is trained on the first frame image. $\text{RoSTFine}_{\alpha=0,1}$ have statistical significance ($p<.05$, {\it Kruskal-Wallis test}) among the other models except TimeSformer on MSE and JS-divergence. Overall, $\text{RoSTFine}_{\alpha=1}$ obtains $0.082\times 10^{-2}$, $0.174\times 10^{-2}$, $2.43$\% higher on MSE, JS-div. and average of BA.}
\label{existing model}
\vspace{-2mm}
\end{table*}

\subsection{Experimental Setup}
\noindent\textbf{Dataset and Task.}\hspace{1mm}
We use the dataset described in \S\ref{sec:dataset}, and the task is 5-point histogram value regression.

\noindent\textbf{Metrics.}\hspace{1mm}
We use Mean Squared Error (MSE) and JS-divergence between predicted \textit{grade distributions} and ground truth distributions as metrics.
%\red{Additionally, we can assign a class ({\it e.g.}, the most selected grade) into a sperm, so that, we use Balanced Accuracy as a metric for classification tasks due to the long-tail distributed dataset ({\it cf}. \S\ref{sec:dataset}). Therefore we evaluate models by five classification tasks whose classes are $n$-th most selected grades, because the worst and the second most selected grades as well as the most selected one are important for decision support.}
Additionally, we evaluate models on classification task by assigning a class ({\it e.g.}, the most selected grade) into a sperm. We use Balanced Accuracy as a metric due to the long-tail distributed dataset ({\it cf}. \S\ref{sec:dataset}). Specifically, we measure Balanced Accuracies on the classification task to predict the $n$-th most selected grade. The worst or the second most selected grades as well as the most selected one are important for decision support.
% さらに、～のため、分類問題としたときの評価も行う。
%However, We actually should use a metric to consider the pecking order of the grades instead of MSE or JS-divergence to calculate the error between each grade independently.
%For future work, we are interested in exploring what metrics are the best.

\noindent\textbf{Implementation Details.}\hspace{1mm}
%   single いらない->done
We conduct experiments on NVIDIA A5000 24GB Single GPU using the Pytorch library \cite{DBLP:conf/nips/PaszkeGMLBCKLGA19}. We download the pretrained weight of TimeSformer from \url{https://github.com/facebookresearch/TimeSformer}.
In the training process, we optimize the models with Stochastic Gradient Descent (SGD) optimizer with learning rate of 1e-3, momentum of 0.9, and weight decay of 5e-4.
Each model is trained with a batch size of 8 and lasted for 200 or 300 epochs. 
We train and evaluate the models on five-fold cross-validation.
The default settings for RoSTFine are $K=60, L=6$, and $H=8$, where $K$ is the number of patches picked out in PSM, $L$ is the number of FGS and FGT branch attention blocks, and $H$ is the number of heads of the multi-head attention.

\subsection{Comparison with Baselines}\label{sec:baseline}

%First, we confirm the superiority of proposed RoSTFine without the diversity loss ($\text{RoSTFine}_{\alpha=0}$) for the sperm videos. The experiment for the diversity loss is described in \S\ref{sec:diversity loss}.
We compare the performances of our $\text{RoSTFine}_{\alpha=0,1}$ with those of the baselines to confirm its superiority for sperm videos. Where $\alpha$ denotes a weight of the diversity loss ({\it cf}. Eq.\ref{eq:train obj}).
The baselines are R3D~\cite{Hara_2017_ICCV}, R(2+1)D~\cite{Hara_2017_ICCV}, X3D~\cite{Feichtenhofer_2020_CVPR}, I3D~\cite{Carreira_2017_CVPR}, SlowFast~\cite{Feichtenhofer_2019_ICCV}, ViViT~\cite{Arnab_2021_ICCV}, and TimeSformer~\cite{pmlr-v139-bertasius21a}.
The models\footnote{We downloaded the pretrained weights of R3D, R(2+1)D, X3D, I3D and SlowFast from \url{https://github.com/facebookresearch/pytorchvideo} and that of ViViT from \url{https://github.com/mx-mark/VideoTransformer-pytorch}.} used in this experiment are pretrained on Kinetics-400~\cite{Carreira_2017_CVPR}.
Furthermore, we consider VGG16~\cite{Simonyan2014VeryDC}, pretrained on ImageNet~\cite{5206848}, as the lower bound, whose inputs are the first frame image.

First, in Table \ref{existing model}, $\text{RoSTFine}_{\alpha=0}$ outperforms the baselines on MSE, JS-divergence (JS) and most Balanced Accuracy (BA), and has statistical significances ($p<.05$, {\it Kruskal-Wallis test}) among the other models except TimeSformer on MSE and JS. Although we cannot obtain statistical significance between $\text{RoSTFine}_{\alpha=0}$ and TimeSformer, $\text{RoSTFine}_{\alpha=0}$ outperform TimeSformer in four out of five folds in MSE and JS ({\it cf.}\S A.3). Therefore, our RoSTFine model are better than TimeSformer for the sperm videos.
Second, $\text{RoSTFine}_{\alpha=1}$ outperforms the baselines and $\text{RoSTFine}_{\alpha=0}$, and obtains the best results on MSE, JS and most BA, which indicates that the diversity loss is effective.
Then, although R(2+1)D and X3D are trained on videos, and VGG16 is trained on only one frame image, the performances of R(2+1)D and X3D are worse than those of VGG16 in Table \ref{existing model}. One possible reason for this is that 
sperm motion factor is negative for sperm recognition when the model cannot capture sperm motion accurately. This suggests that caution should be exercised when designing sperm recognition models.

%Figure \ref{fig:lern} shows the validation MSE loss curves of the top 5 models in Table \ref{existing model} and $\text{RoSTFine}_{\alpha=1}$. The loss in each epoch is calculated as the mean of the MSE losses of the five folds. We can observe that RoSTFine obtains the lowest loss, and the diversity loss helps learning in later epochs.

%\begin{figure}[h]
%\begin{center}
%\includegraphics[width=7cm]{fig/abc.png}
%\end{center}
%   \caption{Learning curves of the validation losses.}
%\label{fig:lern}
%\end{figure}

%In Figure \ref{fig:lern}, We show the validation MSE loss curves of top 5 models in Table \ref{existing model} and $\text{RoSTFine}_{\alpha=1}$. The loss of each epoch is calculated by mean of the MSE losses of 5 folds. We can observe that RoSTFine spends the most time on optimization due to the complexity of the loss of RoSTFine (Eq \ref{eq:MSE_}), but obtains the lowest loss.

\subsection{Visualizing Attention Map} \label{error analysis}
\begin{figure*}
\centering
\includegraphics[width=14cm]{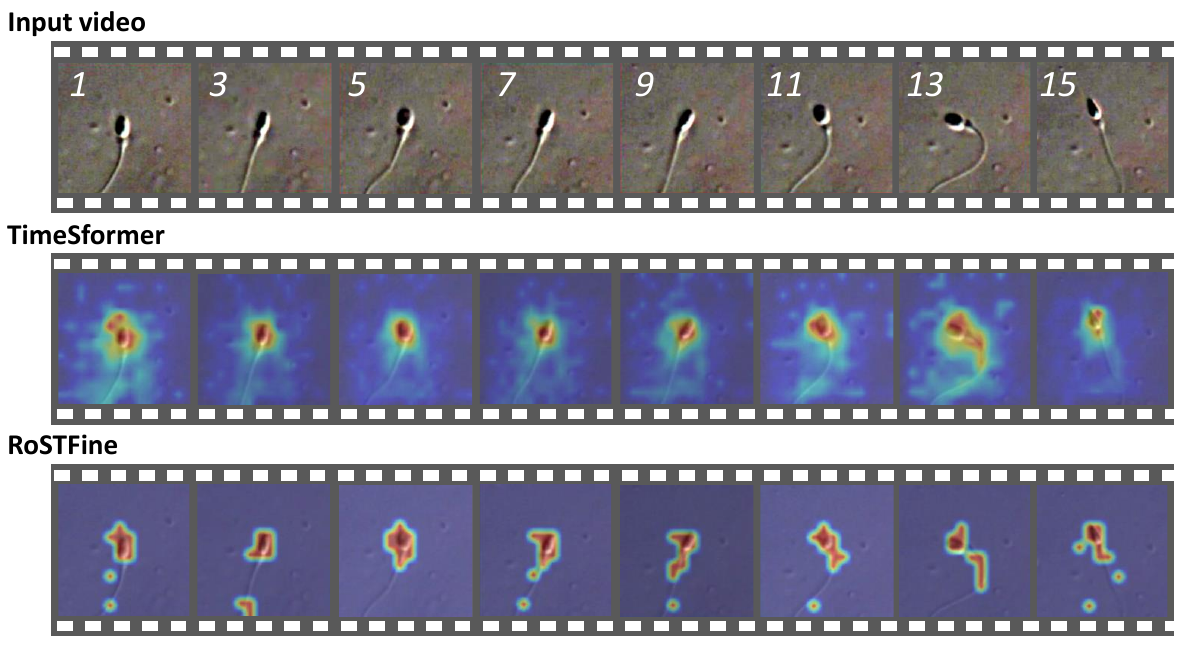}
   \caption{Attention map visualizations of odd number frames of TimeSformer and RoSTFine($K=5$). While TimeSformer attend to a wide area around the sperm, RoSTFine attend strongly to only sperm. RoSTFine can capture particularly the sperm head and neck, and can capture the tip of the tail across frames and the middle of the tail when the tail moves vigorously.}
\label{fig:example}
\end{figure*}

In this section, we analyze the space attention visualizations.
Figure \ref{fig:example} presents the space attention maps of 1st, 3rd, 5th, 7th, 9th, 11th, 13th, and 15th frames obtained by TimeSformer and RoSTFine($K=5$). To visualize attention maps, we use the Attention Rollout scheme~\cite{abnar-zuidema-2020-quantifying}.

In Figure \ref{fig:example}, we observe that TimeSformer focuses on a wide area around the sperm and strongly on the sperm head, whereas RoSTFine focuses hardly on background and only on the sperm, such as the head and neck.
Moreover, we observe that RoSTFine captures the sperm head and neck as well as the sperm tail.
These results suggest two good points of RoSTFine.
First, PSM can reduce redundancy.
Second, RoSTFine can focus on critical parts of the sperm, because the sperm head contains deoxyribonucleic acid, which carries the genetic instructions necessary for reproduction, and the neck of the sperm contains mitochondria, which supply the energy necessary for movement to the tail~\cite{10665-343208}.
Furthermore, The tip of the sperm tail, captured across most frames, and the middle of the sperm tail,  captured when the tail moves vigorously, are necessary for motility.

\subsection{Effectiveness of Diversity Loss} \label{sec:diversity loss}
To confirm the effectiveness of the diversity loss (\S\ref{training and inference}), we compare the task performance of RoSTFine in the range of $\alpha=\{0,0.01,0.05,0.1,0.5,1.0,1.5\}$.

\begin{figure}
\includegraphics[width=8.35cm]{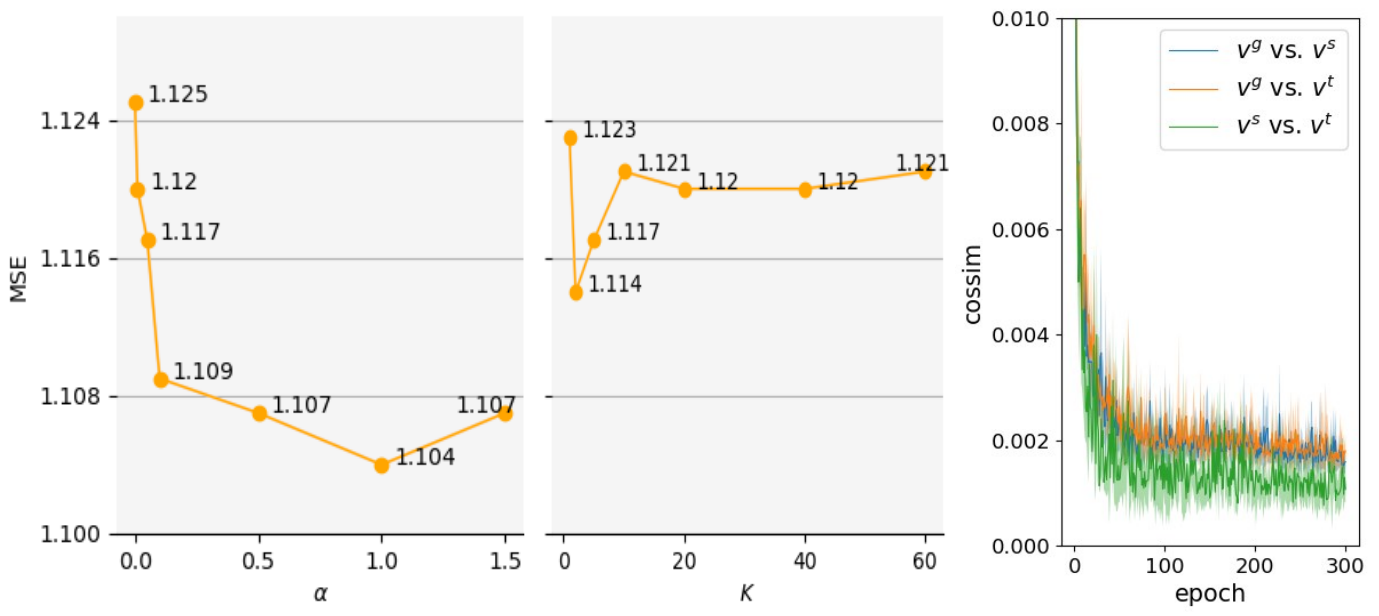}
   \caption{The transition of the MSE loss with the degree $\alpha$ of $\mathcal{L}_{div}$ (left), that with the number $K$ of selected patches (center), and cosine similarity of any features combinations in each epoch (right).}
\label{fig:ka}
\end{figure}

We observe that $\text{RoSTFine}_{\alpha=1.0}$ is 0.017 point higher than $\text{RoSTFine}_{\alpha=0}$ in Table \ref{fig:ka} (left).
Furthermore, we observe that the higher the $\alpha$ value, the higher the performance in the range of $0<\alpha<1$.
The performance is higher than that of the baseline ($\alpha=0$) even when the $\alpha$ value is the smallest ($\alpha=0.01$).
These results indicate that the diversity loss is effective.
Then, Figure \ref{fig:ka} (right) shows the transition of the cosine similarities of any features combinations. We confirm that all cosine similarities are brought close to zero.
We consider that the diversity loss makes $\boldsymbol{v}^g, \boldsymbol{v}^s$, and $\boldsymbol{v}^t$ different from each other to obtain diverse sperm features, which facilitate sperm representation learning.

%In Figure \ref{fig:lern}, we can observe that $\text{RoSTFine}_{\alpha=1.0}$ spends most of the time on optimization owing to the complexity of the diversity loss.

\subsection{Additional Ablations} \label{ablation}
In this section, we investigate RoSTFine in detail by conducting additional experiments to answer the four Research Questions.

\noindent\textbf{RQ1: How many tokens should be selected in PSM?}\\
We are interested in the best number of patches to be selected.
We hypothesize that if the value of $K$ is considerably large, RoSTFine cannot extract the fine-grained features, and if it is significantly small, RoSTFine may miss some important factors, both of which decreases performances.
To verify this hypothesis, we compare the task performances in the range of $K=\{1,2,5,10,20,40,60\}$ and attention maps of RoSTFine in $K=3,5,20$.

\begin{figure}
\centering
\includegraphics[width=8cm]{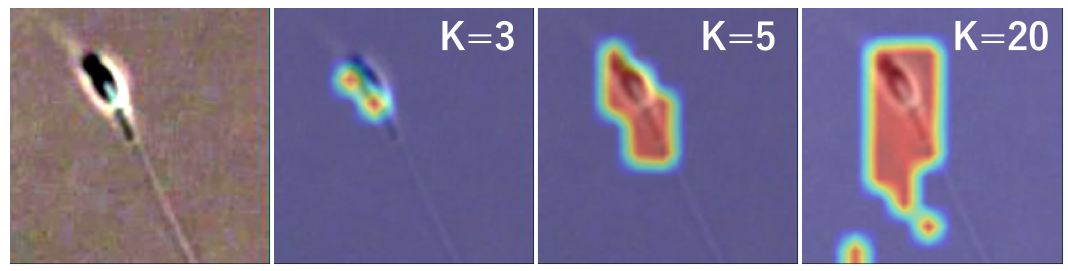}
   \caption{Attention map visualizations of a frame using space attentions of RoSTFine in $K=3,5,20$. We observe that more patches contain background in $K=20$ than that in $K=3,5$.}
\label{fig:kexample}
\end{figure}

Figure \ref{fig:ka} and \ref{fig:kexample} show the task performances and attention maps, respectively.
In Figure \ref{fig:ka} (center), we observe that the performances remain almost the same in range from $K=10$ to $60$, the best is in $K=3$, and the worst is in $K=1$.
In Figure \ref{fig:kexample}, we observe that RoSTFine captures more background in $K=20$ than that in $K=3,5$.
These results indicate that the best number of $K$ is $3$.
When $K=3,5$, selected patches have low redundancy, and RoSTFine focuses almost only on sperm and extracts fine-grained features.
When $10\leq K\leq60$, selected patches are redundant, contain background, and RoSTFine cannot extract the fine-grained features.
When $K=1$, selected patches are so few that they contains insufficient information.
%When $1<K\leq5$, only 40 patches at most are selected, which reduces redundancy in the patches and allows for RoSTFine to learn to capture more fine-grained sperm characteristics.
%When $K=1$, there are considerably few patches and insufficient information, resulting in a sharp drop in performance.

\noindent\textbf{RQ2: Are the features $\boldsymbol{v}^s$ and $\boldsymbol{v}^t$ really effective?}\\
We confirm the effectiveness of the features $\boldsymbol{v}^s$ and $\boldsymbol{v}^t$ generated by FGS and FGT, respectively, by comparing their performances in any combinations of $\boldsymbol{v}^g$, $\boldsymbol{v}^s$, and $\boldsymbol{v}^t$.
We experiment for both training and inference using the same combinations. When using only $\boldsymbol{v}^g$, the model is the same as TimeSformer. 

In Table \ref{combination}, we observe that the performances of only $\boldsymbol{v}^s$ and only $\boldsymbol{v}^t$ is higher than that of only $\boldsymbol{v}^g$, which suggests that $\boldsymbol{v}^s$ and $\boldsymbol{v}^t$ generated by FGS and FGT are effective.
Moreover, we observe that combinations in any other features perform better, and the performance of the combination of all features is the best. This result demonstrates the effectiveness of the feature combination.

\begin{table}[h]
  \begin{center}
    {\small{
\begin{tabular}{cccc}
\toprule
$\boldsymbol{v}^g$ & $\boldsymbol{v}^s$ & $\boldsymbol{v}^t$ & \textbf{MSE}$^{(10^{-2})}$ \\
\midrule
\ding{51}$^{\dag}$ &&& 1.186 \scriptsize{$\pm$ 0.10}\\
&\ding{51}&& 1.138 \scriptsize{$\pm$ 0.10}\\
&&\ding{51}& 1.145 \scriptsize{$\pm$ 0.07}\\
\ding{51}&\ding{51}&& 1.135 \scriptsize{$\pm$ 0.11}\\
\ding{51}&&\ding{51}& 1.139 \scriptsize{$\pm$ 0.11}\\
&\ding{51}&\ding{51}& 1.134 \scriptsize{$\pm$ 0.10}\\
\ding{51}&\ding{51}&\ding{51}& \textbf{1.121} \scriptsize{$\pm$ 0.11}\\
\bottomrule
\end{tabular}
}}
\end{center}
\caption{Performances of any combinations of $\boldsymbol{v}^g$, $\boldsymbol{v}^s$ and $\boldsymbol{v}^t$. $\dag$ denotes the same model as TimeSformer. The performance when using all features is the best.}
\label{combination}
\end{table}

\noindent\textbf{RQ3: How should $\boldsymbol{v}^g$, $\boldsymbol{v}^s$ and $\boldsymbol{v}^t$ be aggregated?}\\
There are some way to aggregate $\boldsymbol{v}^g$, $\boldsymbol{v}^s$ and $\boldsymbol{v}^t$.
We test the following aggregation strategies: (1) concatenating the features (Concat); (2) Summing the features (Sum); (3) Calculating losses separately in training, and the mean of the outputs \{$\boldsymbol{y}^g, \boldsymbol{y}^s, \boldsymbol{y}^t$\} in inference (Ours). 

The results in Table \ref{tab:aggregate} show that our strategy achieves the best performance, which can be attributed to the method of calculating losses separately for multiple features generated on different architectures, thereby enabling the use of more diverse features, such as ensemble learning.
\begin{table}[hbpt]
  \begin{center}
    {\small{
\begin{tabular}{lcc}
\toprule
\textbf{Aggregation type} & \textbf{MSE}$^{(10^{-2})}$ & \textbf{JS divergence}$^{(10^{-2})}$ \\
\midrule
Concat & 1.228 \scriptsize{$\pm 0.01$} & 4.238 \scriptsize{$\pm 0.15$}\\
Sum & 1.197 \scriptsize{$\pm$ 0.10} & 4.153 \scriptsize{$\pm 0.24$}\\
\midrule
Ours & \textbf{1.121} \scriptsize{$\pm$ 0.11} & \textbf{4.145} \scriptsize{$\pm 0.26$}\\
\bottomrule
\end{tabular}
}}
\end{center}
\caption{Performances in three aggregation strategies. (1) concatenating the features (Concat); (2) Summing the features (Sum); (3) Calculating losses separately in training and the mean of the outputs $\boldsymbol{y}^g$, $\boldsymbol{y}^s$ and $\boldsymbol{y}^t$ in inference (Ours). Ours achieves the best performance.}
\label{tab:aggregate}
\end{table}

\noindent\textbf{RQ4: What impact do differences in pretraining data have on sperm recognition?}\\
Pretraining data are an important factor in designing a model because they critically affect the nature of the model.
We investigate the best pretraining data for a sperm recognition task using TimeSformer pretrained on Kinetics~\cite{Carreira_2017_CVPR}, Something-Something-V2 (SSv2)~\cite{goyal2017something}, and HowTo100M (HT100M)~\cite{miech19howto100m}, publicly available at \url{https://github.com/facebookresearch/TimeSformer}. While spatial cues are more important than temporal information in achieving high performance in Kinetics and HT100M, spatial and temporal information are important in SSv2.

\begin{figure}[h]
\begin{center}
\includegraphics[width=6.7cm]{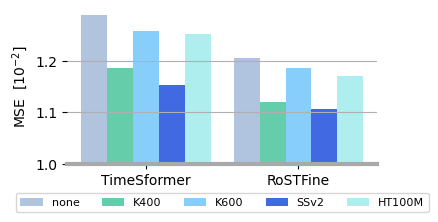}
\end{center}
   \caption{MSE loss of each pretraining dataset. In both of TimeSformer and RoSTFine, SSv2 achieves the best performance.}
\label{fig:ptdata}
\end{figure}

In Figure \ref{fig:ptdata}, both TimeSformer and RoSTFine achieve the lowest loss in SSv2, which suggests that both spatial and temporal sperm characteristics are important. Therefore, a model pretrained on a dataset that requires capturing temporal information is suitable for sperm recognition.

\section{Conclusion}
To assist clinicians to assess sperm and select optimal sperm, in this study, we constructed a sperm video dataset annotated with soft-labels and proposed an automated framework and a neural network, RoSTFine, for sperm assessment. In designing the network, to extract fine-grained and diverse sperm features, Patch Selection Module (PSM) and Role-Separated Branch (RSB) are placed on the head of TimeSformer. PSM filters patches to obtain features that focus on fine-grained sperm characteristics.
RSB can obtain spatial and temporal fine-grained sperm features.
%FGS can obtain spatial fine-grained sperm features by self-attention within each frame, whereas FGT can obtain temporal fine-grained sperm features by self-attention across all frames.
Our experimental sesults showed the superiority of RoSTFine and the effectiveness of PSM and RSB.
We addressed reproduction, an important medical issue in human life but little-studied in computer vision fields, and our study has the potential to make a contribution to human well-being.
%In this study, we used MSE and JS-divergence as metrics, however, we should use metrics to consider the pecking order of the grades instead of MSE or JS-divergence to calculate the error between each grade independently. In the future, we will hear from experts about the relationship between the grades and explore the best metrics.

\paragraph{Limitations.}
(1) We used comprehensive models, in particular publicly available pretrained models, (\S\ref{sec:baseline}) as far as we know, but might have overlooked or updated the other state-of-the-art models.
(2) We cannot confirm the model's robustness to sample preparation and microscope settings, because all samples of our dataset were taken in the same ways and settings. We will develop our work by collecting more samples in various ways and settings.
%------------------------------------------------------------------------

{\small
\bibliographystyle{ieee_fullname}
\bibliography{egbib}
}
\newpage
%------------------------------------------------------------------------
\appendix
\section{Appendix}
\subsection{Domain Knowledge}
In this section, we briefly introduce domain knowledge regarding sperm assessment and assisted reproductive technologies, according to the WHO semen analysis manual~\cite{10665-343208}.
\subsubsection{Biological Structure of Sperm}
\begin{figure}[h]
\includegraphics[width=8cm]{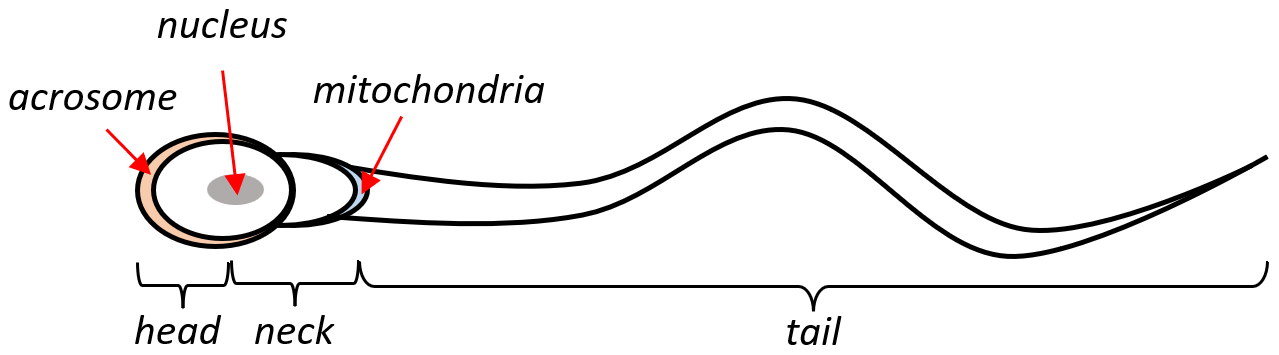}
   \caption{Diagram of human sperm. Normal human sperm has 3 main parts: head, neck and tail.}
\label{fig:sperm}
\end{figure}

Structurally, a normal human sperm has three main parts: the head, neck, and tail, as shown in Figure \ref{fig:sperm}.
The tip of the sperm head contains acrosome that secretes enzymes that are useful for penetration, making the penetration process easier.
The neck contains the mitochondria that supply the tail with energy for movement.
The tail provides the sperm with motility for movement to oocytes for fertilization.

\subsubsection{Assessment as Gold Standard}
We present four key points that experts consider when assessing sperm. \\
\textbf{1. Head Shape:} The head should be smooth, regularly contoured, and generally oval shaped. The ratio of the head width to length should be approximately $3:5$.\\
\textbf{2. Acrosome Size:} The acrosome should comprise 40-70\% of the head area.\\
\textbf{3. No Vacuoles:} The acrosome should contain no vacuoles and no more than two small vacuoles that should not occupy more than 20\% of the sperm head.\\
\textbf{4. Neck:} The neck should be slender, regular, and approximately equal in length to the sperm head.

\subsubsection{Assisted Reproductive Technologies (ARTs)}
ARTs are primarily used to address infertility. ARTs includes in-vitro-fertilization (IVF), intracytoplasmic sperm injection (ICSI), cryopreservation of gametes or embryos, and the use of fertility drugs.
See the WHO Semen Analysis Manual~\cite{10665-343208} for details on ARTs.

\subsection{More Details on Our Paper}
\subsubsection{Samples of the Dataset}
\begin{figure}[h]
\includegraphics[width=8cm]{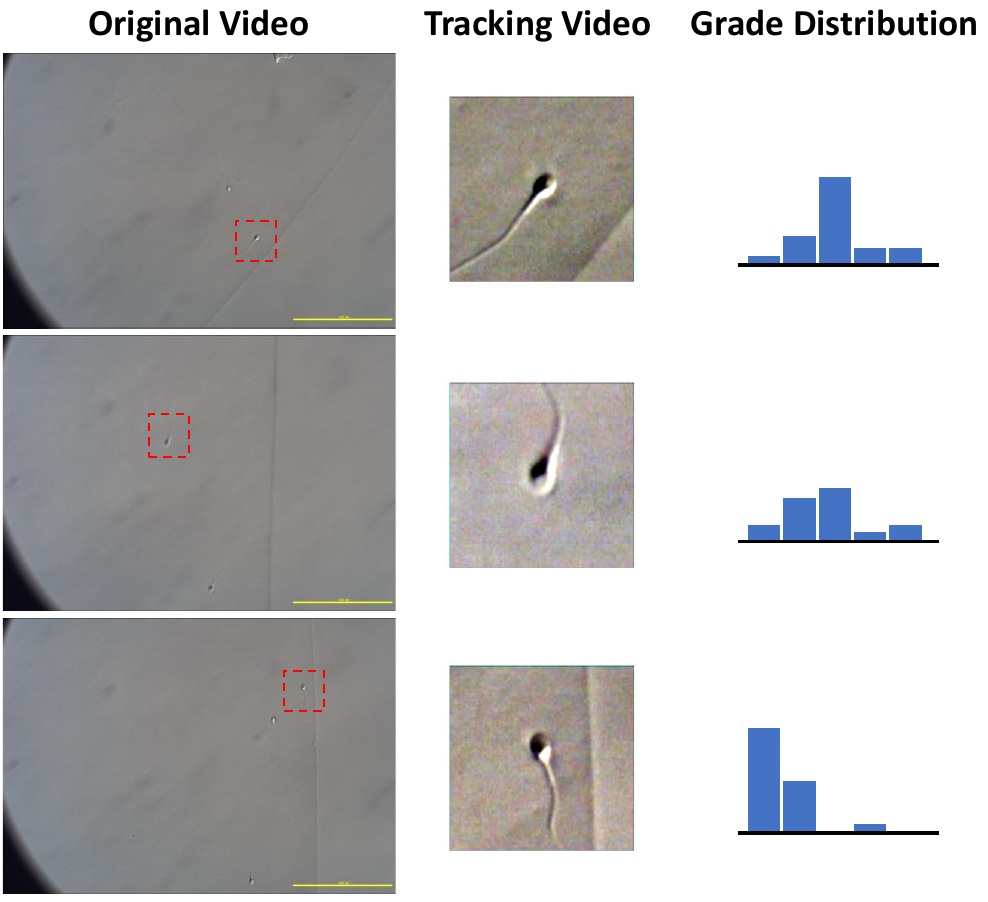}
   \caption{Some samples of the dataset: Original videos (left), Trancking Video (center) and Grade Distribution (right).}
\label{fig:data_sample}
\end{figure}

Some samples of the constructed data are shown in Figure \ref{fig:data_sample}. Original videos (left) are videos taken from a microscope and red regions surround target sperms.

\begin{figure*}[t]
\centering
\includegraphics[width=17.5cm]{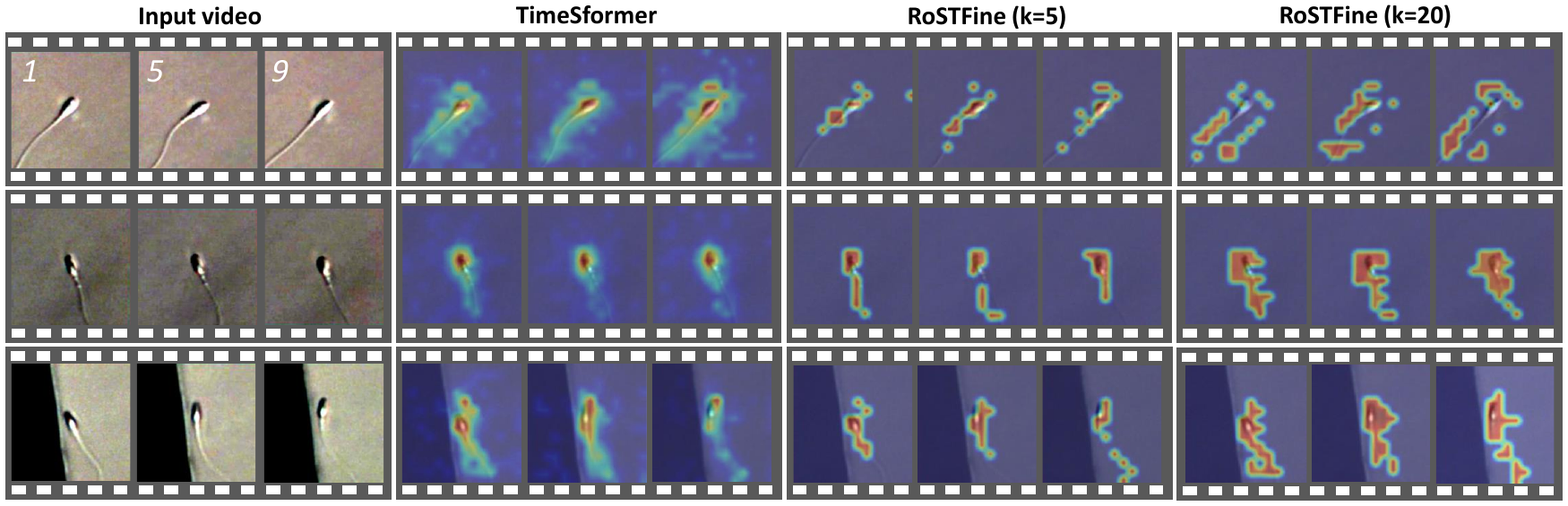}
   \caption{Visualization of attention maps of TimeSformer and RoSTFine ($K=5,20$). }
\label{fig:supp_example}
\end{figure*}

\subsection{TimeSformer v.s. RoSTFine}
Table \ref{vs timesformer} shows performances on each fold.
$\text{RoSTFine}_{\alpha=0}$ outperforms TimeSformer in 4 folds out of 5-fold cross validation in MSE and JS-divergence.
\begin{table}[h]
  \begin{center}
    {\small{
\begin{tabular}{ccccc}
\toprule
\multirow{2}{*}{\textbf{Fold}} &\multicolumn{2}{c}{\textbf{TimeSformer}} &\multicolumn{2}{c}{$\text{RoSTFine}_{\alpha=0}$}\\ \cmidrule(lr){2-3}\cmidrule(lr){4-5}
& MSE$^{(10^{-2})}$ & JS$^{(10^{-2})}$ & MSE$^{(10^{-2})}$ & JS$^{(10^{-2})}$\\
\midrule
1th & 1.138 & \textbf{4.097} & \textbf{1.090}& 4.184\\
2th & 1.112 & 4.293 & \textbf{1.067}& \textbf{4.158}\\
3th & 1.371 & 4.560 & \textbf{1.340}& \textbf{4.558}\\
4th & 1.190 & 4.356 & \textbf{1.073}& \textbf{4.166}\\
5th & \textbf{1.032} & 4.111& \textbf{1.032} & \textbf{3.907}\\
\bottomrule
\end{tabular}
}}
\end{center}
\caption{Each fold performance in TimeSformer and RoSTFine. RoSTFine outperforms TimeSformer in 4 folds out of 5-fold cross validation in the MSE and JS-divergence.}
\label{vs timesformer}
\end{table}
\subsection{Visualizing Attention Map}
In this section, we present the attention maps of TimeSformer and RoSTFine in addition to those in \S5.3.
In Figure \ref{fig:supp_example}, we present space attention visualizations of 1st, 5th and 9th frames obtained using TimeSformer and RoSTFine($K=5,20$). The visualization method is the same as that described in \S5.3.

In the case of $K=5$, we observe that PSM enables the reduction of redundancy and attends more strongly to the sperm in Figure\ref{fig:supp_example}.
Although TimeSformer's attentions are vaguely oriented around the sperm, RoSTFine's attentions are strongly oriented only to the sperm head, neck and tail of the sperm.
This results are the same as those of \S5.3.
However, in the case of $K=20$, RoSTFine captures the sperm as well as a little background.

\end{document}